\documentclass[]{article} 
\usepackage{nips12submit_e,times}
\usepackage{graphicx} 
\usepackage{array} 
\usepackage{mdwlist} 
\usepackage{url}
\usepackage{amsmath,amsfonts,amssymb}

\title{Auto-pooling: Learning to Improve Invariance of Image Features from Image Sequences}

\author{
Sainbayar Sukhbaatar \\
Dept. of Mathematical Informatics, IST \\
The University of Tokyo \\
Tokyo 116-8656, Japan \\
\texttt{sainaa@sat.t.u-tokyo.ac.jp} \\
\And
Takaki Makino\quad\quad Kazuyuki Aihara \\
Institute of Industrial Science \\
The University of Tokyo \\
Tokyo 153-8505, Japan \\
\texttt{\{mak,aihara\}@sat.t.u-tokyo.ac.jp} \\
}

\nipsfinalcopy 

\begin{document}

\maketitle

\begin{abstract}
Learning invariant representations from images is one of the hardest challenges facing computer vision. Spatial pooling is widely used to create invariance to spatial shifting, but it is restricted to convolutional models. In this paper, we propose a novel pooling method that can learn soft clustering of features from image sequences. It is trained to improve the temporal coherence of features, while keeping the information loss at minimum. Our method does not use spatial information, so it can be used with non-convolutional models too. Experiments on images extracted from natural videos showed that our method can cluster similar features together. When trained by convolutional features, auto-pooling outperformed traditional spatial pooling on an image classification task, even though it does not use the spatial topology of features.
\end{abstract}

\section{Introduction}
The main difficulty of object recognition is that the appearance of an object can change in complex ways. To build a robust computer vision, one needs representations that are invariant to various transformations. The concept of invariant features dates back to Hubel and Wiesel's seminal work \cite{hubel1962receptive}, in which cells in a cat's visual cortex are studied. They found two types of cells: simple cells that responded to a specific pattern at a specific location, and complex cells that showed more invariance to location and orientation.

Inspired by simple and complex cells, the spatial pooling step is introduced to computer vision architectures along with the convolution step \cite{fukushima1980neocognitron,lazebnik2006beyond,lecun1998gradient}. In the convolution step, the same feature is applied to different locations. Then in the pooling step, responses from nearby locations are pooled together (typically with a sum or a max operation) to create invariance to small spatial shifting. However, spatial pooling only works with convolutional models. Also, spatial pooling only improves the invariance to spatial shifting. 

An ideal pooling should make features invariant to major types of transformations that appear in the nature. For example, to distinguish people from other objects, one need a representation that is invariant to various transformations of the human body. The complexity of such transformations creates the necessity of more adaptive pooling that does not rely on manual fixed configurations. One promising way to obtain such an adaptive pooling is an unsupervised learning approach.

In recent years, more adaptive spatial pooling methods have been proposed. Jia \textit{et al.}~\cite{jia2012beyond} showed that it is possible to learn custom pooling regions specialized for a given classification task. The training starts with many pooling region candidates, but only a few of them are used in the final classification. This selection of pooling regions is achieved by supervised learning incorporated into the training of a classifier. Although this method learns pooling regions from data, it is still restricted to spatial shifting. Further, it is not suited for deep learning, where lower layers are trained in an unsupervised way.

Another method that improves spatial pooling is proposed by Coates and Ng~\cite{coates2011selecting}, in which local features are clustered by their similarity. A similarity metric between features is defined from their energy correlations. Then, nearby features from the same cluster are pooled together to create rotation invariance. However, the invariance to spatial shifting is still achieved through the same spatial pooling, which restricts this model to convolutional models.

Beside from spatial pooling, there are methods~\cite{hyvarinen2001topographic,kavukcuoglu2009learning,osindero2006topographic} that create invariance by placing features on a two-dimensional topographic map. During training, nearby features are constrained to be similar to each other. Then, invariant representations are achieved by clustering features in a local neighborhood. However, those methods fix clustering manually, which restricts clusters from having adaptive sizes that depend on the nature of their features. Also, we cannot guarantee that the optimal feature clustering can be mapped into two-dimensional space. For example, edge detectors have at least four dimensions of variation, so an ideal clustering can be achieved by placing edge detectors in a four-dimensional space and grouping nearby features. It will be difficult to approximate such a clustering  with a two-dimensional map. Moreover, those methods cannot be used with features already learned by another model.

Slowness has been used in many methods as a criterion for invariant features~\cite{berkes2005slow,mobahi2009deep,cadieu2009learning,gregor2010emergence}. The intuition is that if a feature is invariant to various transformations, then its activation should change slowly when presented with an image sequence containing those transformations. Mobahi et al.~\cite{mobahi2009deep} incorporated unsupervised slowness learning with supervised back-propagation learning, which improved the classification rate. However, our focus is a simple unsupervised method that can make features invariant without changing them, so it can easily replace and improve spatial pooling in any application.

In this paper, we propose \textit{auto-pooling}, a novel pooling method that learns soft clustering of features from image sequences in an unsupervised way. Our method improves the invariance of features using temporal coherence of image sequences. Two consecutive frames are likely to contain the same object, so auto-pooling minimizes the distance between their pooled representations. At the same time, the information loss due to pooling is also minimized. This is done by minimizing the reconstruction error in the same way as auto-encoders \cite{marc2006efficient,vincent2008extracting}. Through experiments, we show that our method can pool similar features and increase accuracy of an image classification task.

\begin{figure}[t]
\centering
\includegraphics{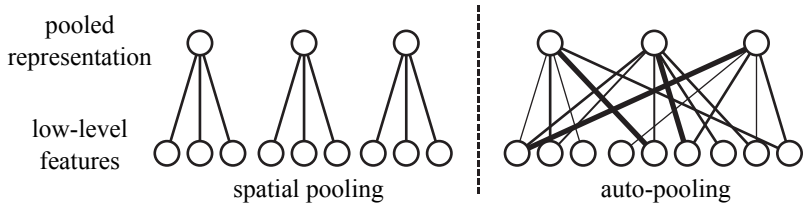}
\caption{In spatial pooling, low-level features are ``hard'' clustered by their spatial organization. Auto-pooling, on the other hand, learns ``soft'' clustering of low-level features from image sequences.}
\label{fig:sp_ap}
\end{figure}

There are several advantages in auto-pooling over traditional spatial pooling (see Figure~\ref{fig:sp_ap}). First, it produces invariance to all types of transformations that present in natural videos. Second, auto-pooling is a more biologically plausible model for complex cells because its parameters are learned from image sequences rather than being manually defined. Third, auto-pooling can be used with non-convolutional models because it does not use spatial information. 

\section{Auto-pooling}
\begin{figure}[t]
\centering
\includegraphics[scale=1]{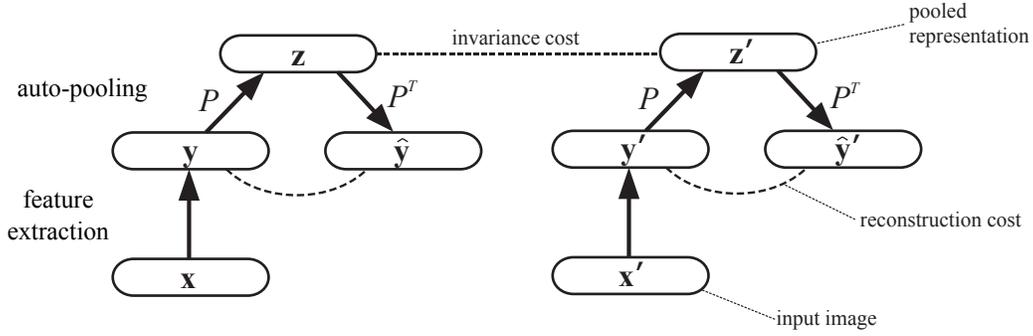} 
\caption{Structure of an auto-pooling model}
\end{figure}

Auto-pooling is a pooling method that learns transformations appeared in image sequences. It is trained by image sequences in an unsupervised way to make features more invariant. The goal of training is to cluster similar features together, so that small transformations would not affect pooled representations. Two features are considered similar if they are traversable by a small transformation such as shift or rotation. We use natural videos as the resource for learning similarity between features, because they are rich in various complex transformations. Moreover, image sequences are available to animals and humans as early as their birth, so it is biologically plausible to use them in learning of complex-cell-like invariant features.

We believe that there are two desirable properties in good pooling methods. The first property is that if two images show the same object, then their pooled representations should be the same. Auto-pooling tries to meet this invariance property by minimizing the distance between pooled representations of consecutive frames, which are likely to contain the same object. The second desirable property is that the information loss due to pooling should be minimal. This is the same as maximizing the cross entropy between inputs and their pooled representations. This entropy property could be obtained by minimizing the error between inputs and reconstructions from their pooled representations.

Instead of image sequences, it is convenient to use image pairs taken from consecutive video frames as training data. Such image pairs can be written as
\[ X = \{ 
\mathbf{x}_{1}, 
\mathbf{x}'_{1}, 
\mathbf{x}_{2},
\mathbf{x}'_{2},
..., 
\mathbf{x}_{N},
\mathbf{x}'_{N}
\}.
\]
If an object is present in image $\mathbf{x}_i$, then it is likely to be present in $\mathbf{x}'_i$ too, because frames $\mathbf{x}_i$ and $\mathbf{x}'_i$ have very small time difference (only 33ms for videos with 30 frames per second). Let us assume that the low-level feature extraction is done by 
 \[ \mathbf{y}_{i} = f(\mathbf{x}_{i}) , \quad
 \mathbf{y}'_{i} = f(\mathbf{x}'_{i}) .\]
Here, $f$ can be any function as long as $\mathbf{y}_{i}$, $\mathbf{y}'_{i}$ are non-negative.

In auto-pooling, clustering of features is parameterized by a pooling matrix $P$. We require all elements of $P$ to be non-negative because they represent the associations between features and clusters. $P_{ij}=0$ means that $j$-th feature does not belong to $i$-th cluster. On the other hand, large $P_{ij}$ indicates that $i$-th cluster contains $j$-th feature. Then, pooling is done by a simple matrix multiplication, that is
 \[ \mathbf{z}_i = P \mathbf{y}_i , \quad
  \mathbf{z}'_i = P \mathbf{y}'_i .\]
If the dimension of feature vectors $\mathbf{y}_i, \mathbf{y}'_i$ is $M$, and the dimension of pooled representations $\mathbf{z}_i, \mathbf{z}'_i$ is $K$, then $P$ is a $K\times N$ matrix.
While in spatial pooling, elements of $P$ are fixed to 0 or 1 based on the topology of feature maps, auto-pooling generalizes it by allowing $P_{ij}$ to take any non-negative value.

Our main goal is to learn pooling parameters $P_{ij}$ from data, without using the explicit spatial information. Training of auto-pooling is driven by two cost functions. The first cost function
  \[ J_1 = \frac{1}{N}\sum_{i=1}^N
 \frac{1}{2} \| \mathbf{z}_i - \mathbf{z}'_i \|_2^2 \]
is for the invariance property, and minimizes the distance between pooled representations $\mathbf{z}_i$ and $\mathbf{z}'_i$. However, there is a trivial solution of $P = 0$ if we use only this cost function. 

The second cost function corresponds to the entropy property, and encourages pooled representations to be more informative of their inputs. Input $\mathbf{y}_i$ and $\mathbf{y}'_i$ are reconstructed from their pooled representations by 
\[ \hat{\mathbf{y}}_i = P^T \mathbf{z}_i , \quad  
\hat{\mathbf{y}}'_i = P^T \mathbf{z}'_i\]
using the same parameters as the pooling step. Then, the reconstruction error is minimized by the cost function of
\[ J_2 = \frac{1}{N}\sum_{i=1}^N
 \frac{1}{2} ( \| \mathbf{y}_i - \hat{\mathbf{y}}_i \|_2^2 + \| \mathbf{y}'_i - \hat{\mathbf{y}}'_i \|_2^2 ) .
\]
This prevents auto-pooling from throwing away too much information for the sake of invariance. 

Auto-pooling is similar to auto-encoders, which are used in feature learning. In auto-pooling, the input is reconstructed in the same way as auto-encoders. Also, the reconstruction cost function $J_2$ is exactly same as the cost function of auto-encoders. However, there are several important differences between them. First, parameters of auto-pooling are restricted to non-negative values. Second, activation functions of auto-pooling are linear and have no biases. Third, auto-pooling has an additional cost function for temporal coherence.

The final cost function of auto-pooling is
 \[ J = \lambda J_1 + J_2 ,\] 
where the parameter $\lambda \ge 0$ controls the weight of the invariance cost function. Larger $\lambda$ will make features more invariant by discarding more information. The training is done by minimizing the cost function with a simple gradient descent algorithm.

\subsection{Invariance Score}
\label{sec:tc_score}
For evaluating our pooling model, we define a score for measuring invariance of features (a similar score is also introduced in~\cite{goodfellow2009measuring}). A simple measurement of feature's invariance is its average activation change between two neighboring frames, which is
\[ G = \frac{1}{N} \sum_{n=1}^N 
\| g(\mathbf{x}_n) - g(\mathbf{x}'_n) \|_2 .\]
Here, $g(\mathbf{x}):=f(\mathbf{x})$ if we are measuring invariance of raw features, and $g(\mathbf{x}) := P f(\mathbf{x})$ if we are measuring invariance of pooled representations.

For invariant features, $G$ should be small. However, features can cheat by simply making its activation constant to reduce $G$ , which is obviously not useful. An ideal invariant feature should take the same value only if the stimuli are from consecutive frames. For frames chosen from random timings, an invariant feature should have different activities because it is likely that the inputs contain different objects. Therefore, the average distance between two random frames
\[ H = \frac{1}{N} \sum_{n=1}^N 
\| g(\mathbf{x}_n) - g(\mathbf{x}'_{\sigma(n)}) \|_2 \]
should be large for invariant features. Here $\sigma$ is a random permutation of $\{1,2,...,N\}$. 
The invariance score is defined as
\begin{equation}
F = \frac{H}{G},
\label{eq:tc_score}
\end{equation}
which will be large only if a feature is truly invariant.

\section{Experiments}
We will show the effectiveness of our method with two types of experiments. In the first experiment, we train an auto-pooling model with non-convolutional features. The goal of this experiment is to see whether similar features are being pooled together. We also measured the invariance score of features before and after pooling. In the second experiment, we compared our method with traditional spatial pooling on an image classification task.

\subsection{Clustering of Image Features}
\begin{figure}[h]
\centering
\includegraphics[width=12cm]{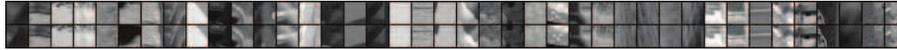}
\caption{Patch pairs (each column) extracted from natural videos}
\label{fig:vimeo16}
\end{figure}

The goal of this set of experiments is to analyze feature clusters learned by auto-pooling. We prepared a dataset of $16\times16$ gray-scale patch pairs from natural videos\footnote{We used 44 short videos. All videos are obtained from \url{http://www.vimeo.com} and had the Creative Commons license. Although we tried to include videos with the same objects as CIFAR10 (a labeled image dataset used in the next experiment), image patches extracted from the videos were qualitatively different than images from CIFAR10. Even if a video contains a desired object, not all frames show the object. Also, most patches only included a part of an object, because the patch size was much smaller than the frame size, }. Patch pairs are extracted from random locations of consecutive frames. Some of the patch pairs are shown in Figure~\ref{fig:vimeo16}.

\begin{figure}[th]
\centering
\includegraphics[width=13cm]{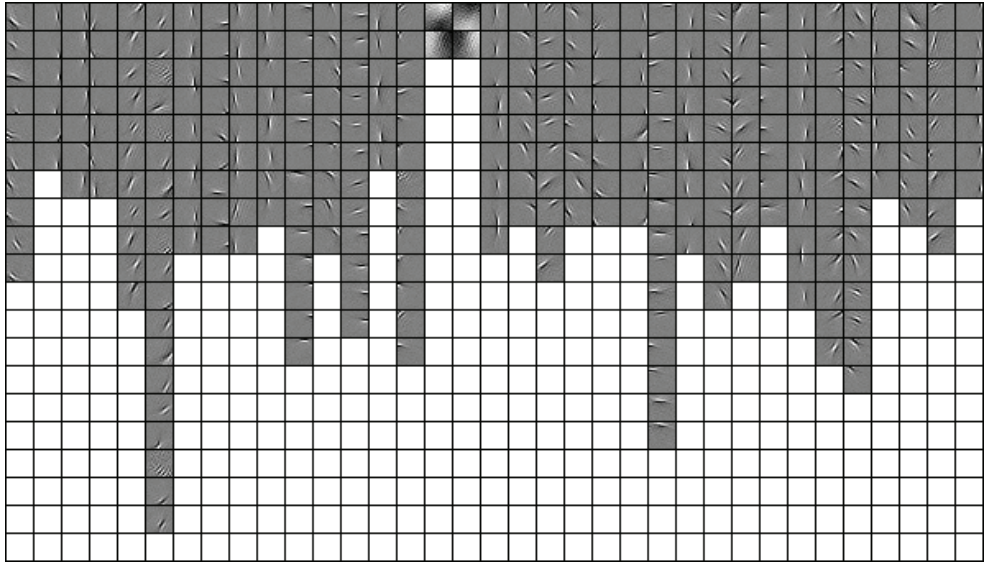}
\caption{Clusters (shown in each column) of features learned from the patch pair dataset}
\label{fig:pooled_features}
\end{figure}

We used a sparse auto-encoder to learn 400 features from patches. Then, we trained an auto-pooling model on those features. Since auto-pooling performs soft clustering, it is hard to visualize the clustering result. For simplicity, we used a small threshold to show some of the learned clusters in Figure~\ref{fig:pooled_features}, where each column represents a single cluster. For $i$-th cluster, we showed features with $P_{ij} > \varepsilon$. It is evident that similar features are clustered together. Also, one can see that the size of a cluster vary depending on the nature of its features. 

\begin{figure}[th]
\centering
\includegraphics[scale=1]{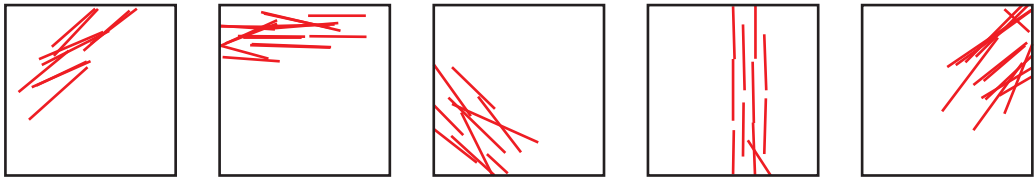}
\caption{Diversity of edge detectors in a single cluster}
\label{fig:pooled_edges}
\end{figure}

To display clusters more clearly, some clusters of edge detectors are shown in more detail in Figure~\ref{fig:pooled_edges}, in which edge detectors are replaced by corresponding thin lines. This allows us to see the diversity of edge detectors inside each cluster. The important thing is that there is variance in orientations as well as in locations, which means that auto-pooling can create representations invariant to small rotations.

\begin{figure}[thbp]
\centering
\includegraphics[width=10cm]{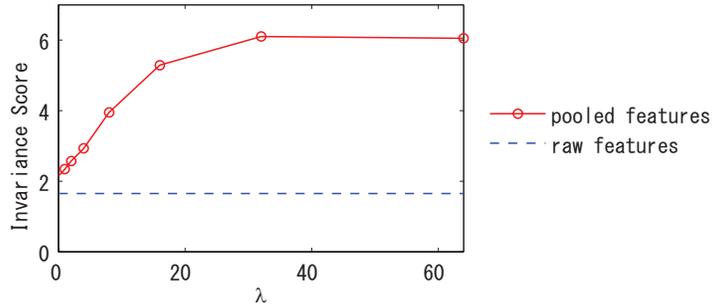}
\caption{Invariance scores before and after pooling}
\label{fig:tc_score}
\end{figure}

Next, we analyzed the effect of pooling on image features using our invariance score. Figure~\ref{fig:tc_score} shows invariance scores measured at various values of $\lambda$, which controls the weight of the invariance cost. The invariance score is significantly improved by the pooling, especially for large $\lambda$ values. It is not surprising because larger $\lambda$ puts more importance on the invariance cost, thus makes pooled representations less likely to change between consecutive frames. As a result, $G$ in equation~\ref{eq:tc_score} becomes smaller and increases the invariance cost.

At the same time, however, increase of $\lambda$ diminishes the role of the reconstruction cost, which was preventing the loss of too much information. Too large $\lambda$ makes pooled representations over-invariant, having constant values all the time. This results in a small $H$ and decreases the invariance cost. This side effect of large $\lambda$ can be observed from Figure~\ref{fig:tc_score}, where the invariance score stopped its increase at large $\lambda$.

\subsection{Image Classification}
\begin{figure}[ht]
\centering
\includegraphics[trim =24.5mm 10mm 21.5mm 10mm, clip, height=13cm, angle=-90]{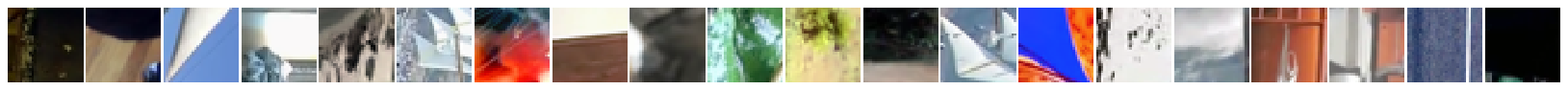}
\\
\includegraphics[trim =24.5mm 10mm 21.5mm 10mm, clip, height=13cm, angle=-90]{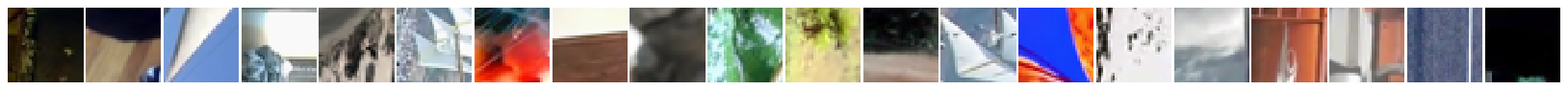}
\caption{Samples from the patch pair dataset used in this experiment}
\label{fig:vimeoColor}
\end{figure}

Next, we tested the effectiveness of our pooling method by applying it to an image classification task. We used two datasets. For image classification, we used CIFAR10 dataset~\cite{krizhevsky2009learning}, which contains 50 thousand labeled images from ten categories. All images were 32$\times$32 pixels in size, and had three-color channels. For training of an auto-pooling model, we prepared a patch pair dataset in the same way as the previous experiment, except patches were 32$\times$32 color images to match CIFAR10 images. Some samples from the patch pair dataset are shown in Figure~\ref{fig:vimeoColor}. 

\begin{figure}[h]
\centering
\includegraphics[width=12cm]{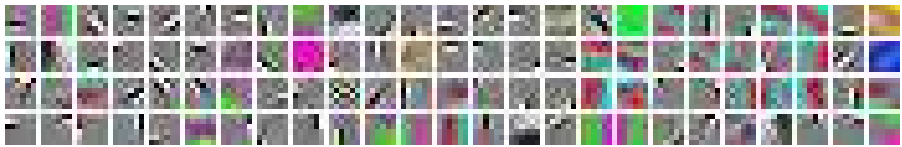}
\caption{Features learned by a sparse auto-encoder from small image patches}
\label{fig:features6x6}
\end{figure}

In the feature extraction step, we used a convolutional model. We trained a sparse auto-encoder with 100 hidden units on 6$\times$6 small patches extracted from CIFAR10 images. Learned local features are shown in Figure~\ref{fig:features6x6}. Then, those $6\times6$ local features are duplicated to all possible locations of 32$\times$32 images, resulting in 100 feature maps of $27\times27$. 

\begin{figure}[h]
\centering
\begin{tabular}{cc}
\includegraphics[width=6cm]{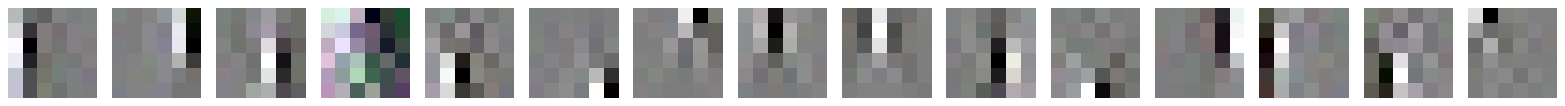} & \includegraphics[width=7cm]{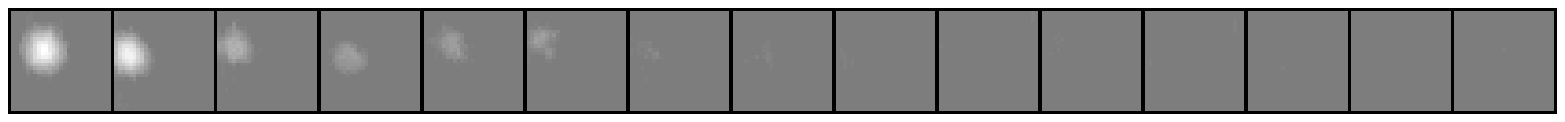} \\
\includegraphics[width=6cm]{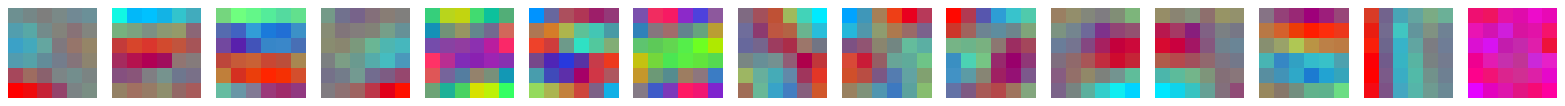} & \includegraphics[width=7cm]{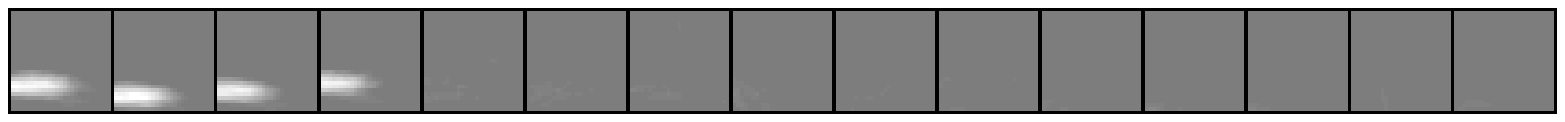} \\
\includegraphics[width=6cm]{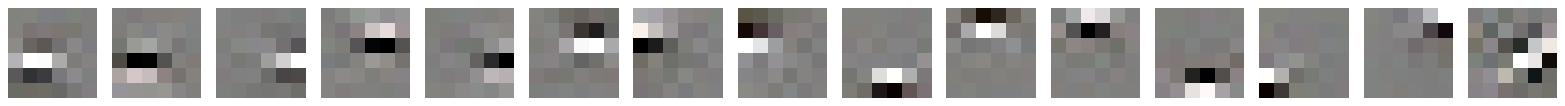} & \includegraphics[width=7cm]{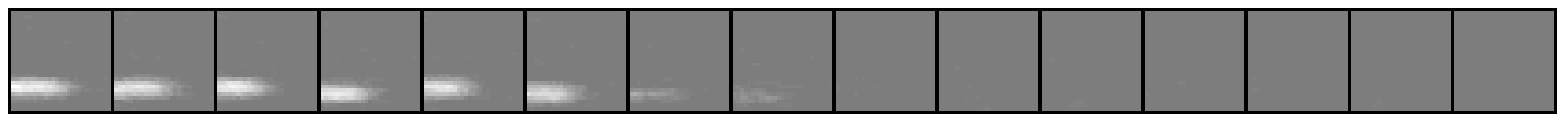} \\
\includegraphics[width=6cm]{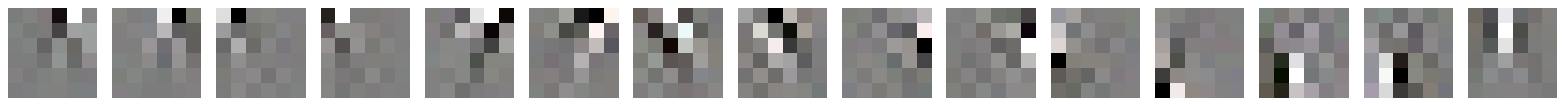} & \includegraphics[width=7cm]{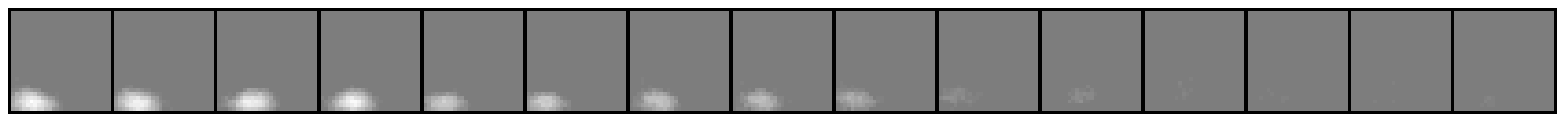} \\
\includegraphics[width=6cm]{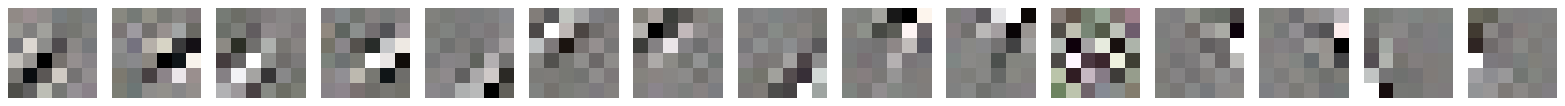} & \includegraphics[width=7cm]{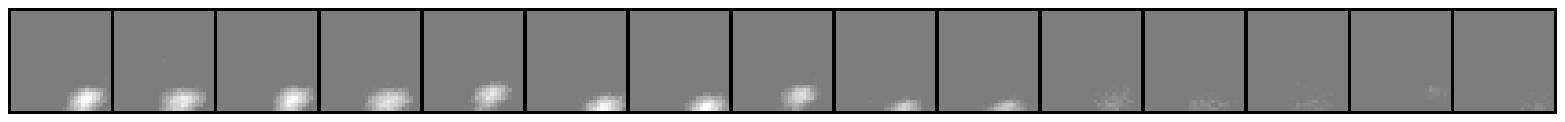}
\end{tabular}
\caption{Feature clusters learned by an auto-pooling model. The right side shows pooled regions on feature maps, and the left side shows corresponding local features.}
\label{fig:pooled_regions_full}
\end{figure}

The convolutional feature extraction produced very large (exceeding 100 gigabytes) training data for auto-pooling. Luckily, the training took only few hours because we implemented our algorithm on a graphic card (Tesla K20c) using CUDAMat library~\cite{mnih2009cudamat}. Some of the learned feature clusters are visualized in Figure~\ref{fig:pooled_regions_full}. For each cluster, we showed 15 feature maps with the largest pooling area (i.e. $\max_k (\sum_{j\in S_k} P_{ij})$, where $S_k$ is the set of features in $k$-th feature map). $P_{ij}=0$ is shown in gray and large $P_{ij}$ is shown in white. Local features corresponding to the feature maps are also shown in Figure~\ref{fig:pooled_regions_full}.

Unlike spatial pooling, each cluster learned by auto-pooling extended to multiple feature maps. Pooling regions (i.e., white areas in Figure~\ref{fig:pooled_regions_full}) of those maps usually have the same continues spatial distribution, which will create spatial invariance in the same way as spatial pooling. If we observe those pooling regions carefully, however, we can see the small variance in their locations. This location variance is inversely related to the location variance of corresponding local features. For example, if there is a edge detector in the lower part of a $6\times6$ local feature, corresponding pooling regions will have upper position in $27\times27$ feature maps.

Beside from pooling by spatial areas, auto-pooling also succeeded in clustering similar local features. In some clusters, edge detectors of similar orientations are grouped together. This will make pooled representations invariant to small rotations, which is a clear advantage over traditional spatial pooling. In addition, clustering of local features only differing in their locations will reduce the redundancy created by convolutional feature extraction.

\begin{table}[b]
\caption{Classification accuracy on CIFAR10 (AP=auto-pooling, SP=spatial pooling)}
\label{tbl:results}
\begin{center}
\begin{tabular}{lll}
 & \textbf{Accuracy} & \textbf{Accuracy} \\ 
\textbf{Pooling methods}&  \textbf{(full)} & \textbf{(small)} \\ \hline \noalign{\smallskip}
AP (400 clusters) &  64.6\% & 61.6\% \\ 
AP (800 clusters) &  67.2\% & 62.9\% \\ 
AP (1300 clusters) &  69.0\% & 63.8\% \\ 
AP (1600 clusters) &  69.4\% & 64.3\% \\ 
AP (2000 clusters) &  69.7\% & 65.0\% \\ 
AP (2500 clusters) &  69.3\% & 63.5\% \\ \noalign{\smallskip} \hline \noalign{\smallskip}
SP ($2\times2$) &  63.8\% & 60.7\% \\ 
SP ($3\times3$) &  68.2\% & 61.5\% \\ 
SP ($2\times2+3\times3$) &  68.2\% & 61.2\% \\ 
SP ($4\times4$) &  68.4\% & 58.7\% \\ 
SP ($2\times2+3\times3+4\times4$) &  68.2\% & 57.6\% 
\end{tabular} 
\end{center}
\end{table} 

\begin{figure}
\centering
\includegraphics[height=7cm]{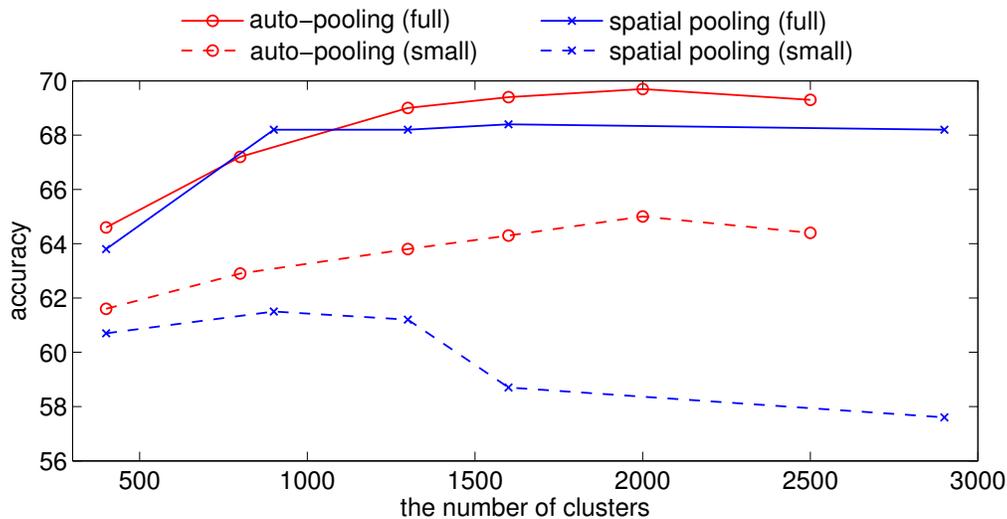}
\caption{Classification accuracies are plotted as functions of the number of features}
\label{fig:results}
\end{figure}

We compared our pooling method with traditional spatial pooling on a classification task, in which a supervised classifier is trained by pooled representations of labeled images. For auto-pooling, we varied the number of clusters from 400 to 2500. For spatial pooling, we can only change the grid size. However, it is possible to use multiple spatial pooling at once~\cite{lazebnik2006beyond} to produce better results. We denote a spatial pooling that used $2\times2$ and $3\times3$ grids by $2\times2+3\times3$.

In classification, we trained a linear SVM with pooled representations. The results are shown in Table~\ref{tbl:results}. We trained the classifier with two training data: a full data with 5000 examples per class, and a smaller one with 1000 examples per class. Since the number of features is an important factor in classification, we plotted the accuracy of the two pooling methods against the number of clusters in Figure~\ref{fig:results}. Auto-pooling outperformed traditional spatial pooling for most of the time. Especially for small training data, the difference between the two pooling methods was substantial. This indicates that auto-pooling is better at generalization, which is the main goal of invariant features. The spatial pooling, on other hand, shows the sign of over-fitting when its pooling regions are increased.

\section{Conclusions}
In this paper, we introduced auto-pooling, a novel pooling method that can generalize traditional spatial pooling to transformations other than spatial shifting. Auto-pooling tries to make features more temporally coherent, having slow changing activation when presented with a continues image sequence. The information loss due to pooling is kept minimum using the same cost function as auto-encoders. The main advantage of our method is that it learns to cluster features from data, rather than relying on manual heuristic spatial divisions. Therefore, auto-pooling is a more biologically plausible model for complex cells.

When trained by image pairs extracted from natural videos, auto-pooling successfully clustered similar features together. We showed that such clustering could significantly improve the invariance of features. Also, our pooling model was more effective than traditional spatial pooling when it was used in a real-world classification task, where spatial pooling had the advantage of using spatial information of features.

In our experiments, the advantage of auto-pooling over spatial pooling was mainly restricted to learning of rotation invariance. This is because auto-pooling is applied to low-level features, which were mostly edge detectors with the size. Therefore, the only possible variance beside spatial shifting was rotation. We believe that if we use auto-pooling instead of spatial pooling in deep architectures, we can create invariance to more complex transformations such as three-dimensional rotations and distortions.

\subsubsection*{Acknowledgments}
This research is supported by the Aihara Innovative Mathematical Modelling Project, the Japan Society for the Promotion of Science (JSPS) through the ``Funding Program for World-Leading Innovative R\&D on Science and Technology (FIRST Program),'' initiated by the Council for Science and Technology Policy (CSTP).

\small
\bibliographystyle{plain} 
\bibliography{paper}

\end{document}